\pdfoutput=1

\documentclass[11pt]{article}

\usepackage[preprint]{acl}

\usepackage{times}
\usepackage{latexsym}

\usepackage[T1]{fontenc}

\usepackage[utf8]{inputenc}

\usepackage{microtype}

\usepackage{inconsolata}

\usepackage{graphicx}
\usepackage{algorithm}
\usepackage{amsmath}
\usepackage{amssymb}
\usepackage{algpseudocode}

\usepackage{booktabs}
\usepackage{colortbl}
\usepackage{pifont}
\usepackage{multirow}
\usepackage{makecell}
\usepackage{mathrsfs}
\usepackage{paralist}
\usepackage{enumitem}
\usepackage{cleveref}

\definecolor{lightpurple}{HTML}{E8E8FF}
\definecolor{upgreen}{HTML}{228B22}
\definecolor{downred}{HTML}{FF4500}

\title{
    Full-Step-DPO: Self-Supervised Preference Optimization with Step-wise Rewards for Mathematical Reasoning
}

\author{Huimin Xu$^1$, Xin Mao$^1$, Feng-Lin Li$^2$, Xiaobao Wu$^1$, Wang Chen$^2$, Wei Zhang$^{3}$, Anh Tuan Luu$^{1}*$\\
  $^1$Nanyang Technological University, Singapore \\
  $^2$Shopee Pte. Ltd, Singapore, $^3$SEA Group, Singapore\\\
  \texttt{\{huimin.xu, xin.mao, anhtuan.luu\}@ntu.edu.sg}, \texttt{xiaobao002@e.ntu.edu.sg} \\ 
  \texttt{\{fenglin.li, chen.wang\}@shopee.com}, \texttt{terry.zhang@sea.com}}

\begin{document}
\maketitle
\begin{abstract}

Direct Preference Optimization (DPO) often struggles with long-chain mathematical reasoning. Existing approaches, such as Step-DPO, typically improve this by focusing on the first erroneous step in the reasoning chain. However, they overlook all other steps and rely heavily on humans or GPT-4 to identify erroneous steps. To address these issues, we propose Full-Step-DPO, a novel DPO framework tailored for mathematical reasoning. Instead of optimizing only the first erroneous step, it leverages step-wise rewards from the entire reasoning chain. This is achieved by training a self-supervised process reward model, which automatically scores each step, providing rewards while avoiding reliance on external signals. Furthermore, we introduce a novel step-wise DPO loss, which dynamically updates gradients based on these step-wise rewards. This endows stronger reasoning capabilities to language models. Extensive evaluations on both in-domain and out-of-domain mathematical reasoning benchmarks across various base language models, demonstrate that Full-Step-DPO achieves superior performance compared to state-of-the-art baselines~\footnote{Our code, data, and models will be made publicly available after the publication of this paper.}.
\end{abstract}

\section{Introduction}
\label{sec:1}

Large Language Models (LLMs) have attracted massive interest due to their remarkable capabilities across various tasks \cite{kaddour2023challenges, song2023restgpt, wang2023voyager, zheng2024judging, wang2023large,pan2024fallacy,wu2024fastopic,wu2024akew,wu2024antileak,wu2024survey}.
However, they commonly encounter difficulties when tackling complex and symbolic multi-step reasoning, particularly in mathematical problem reasoning \cite{lightman2023let, huang2023large}. 
To improve the mathematical reasoning ability, some studies use Direct Preference Optimization (DPO) \cite{dpo} with solution-wise preference data but find its benefit limited \cite{smaug,chatglm,pdpo}.
Recent works attribute this limitation to DPO's inability to perform process supervision and instead builds preference data based on reasoning steps rather than entire solutions \cite{cuhkstep, alistep, nusstep, stepcont}.
For example, Step-DPO \cite{cuhkstep} focuses on optimizing only the first erroneous step in the reasoning chain, demonstrating notable improvements.

\begin{figure}[t]
\centering
\includegraphics[width=\linewidth]{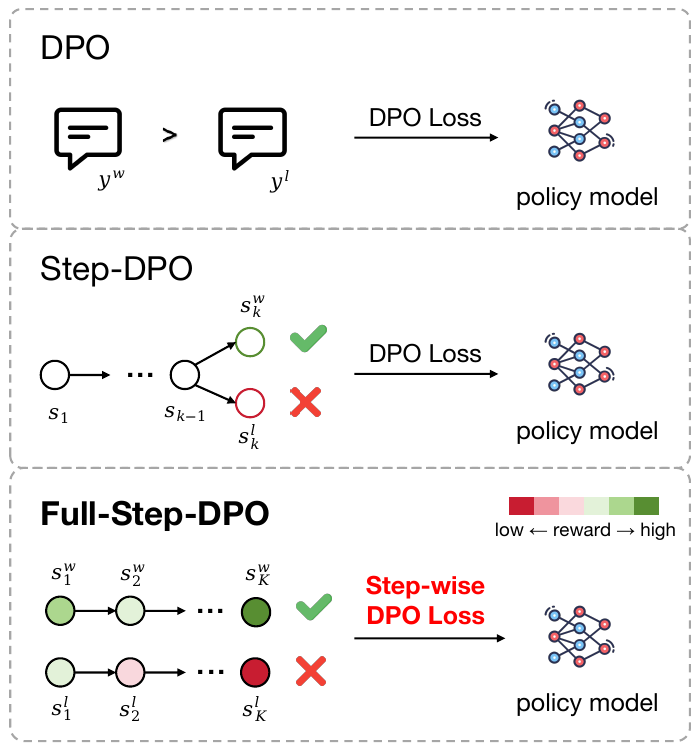}
\caption{
    Comparison between DPO, Step-DPO, and our \textbf{Full-Step-DPO}.
    DPO operates on solution-wise preference data.
    Step-DPO advances to step-wise data but optimizes only a single step.
    Full-Step-DPO optimizes all steps with a novel step-wise DPO loss, effectively enhancing the model's reasoning capability.
}
\label{fig:fig1}
\end{figure}

However, despite their improvements, these existing methods face the following limitations:
\begin{inparaenum}[(1)]
    \item
        Some focus solely on the first erroneous step and ignore all other useful steps in the reasoning chain \cite{cuhkstep}, as shown in \Cref{fig:fig1}.
        As a result, they fail to fully optimize the reasoning chains, leading to suboptimal performance.
    \item
        Their loss function still follows the early vanilla DPO in a solution-wise approach \cite{stepcont, nusstep}.
        Consequently, it cannot directly leverage rewards in a step-wise fashion for learning.
    \item
        They heavily rely on costly and resource-intensive annotations from GPT-4 or humans to detect erroneous steps \cite{cuhkstep, lightman2023let},
        significantly limiting their practicality.
\end{inparaenum}

To address the above limitations, we propose \textbf{Full-Step-DPO}, a novel DPO framework for mathematical reasoning.
As illustrated in \Cref{fig:fig1}, unlike vanilla DPO, which operates solution-wise, or Step-DPO, which focuses solely on the first erroneous step, Full-Step-DPO utilizes each step in the entire reasoning chain and optimizes them using step-wise rewards.
We first train a Process Reward Model (PRM) \cite{lightman2023let,mathshepherd} in a self-supervised way, utilizing data generated by the model itself.
This approach enables the PRM to automatically score each step in the reasoning chain, eliminating the reliance on external annotations such as GPT-4 or humans.
Then, we propose a novel \textbf{Step-wise DPO Loss}, which employs dynamic gradient updates to optimize each step based on its corresponding reward.
This approach shifts the optimization focus from solution-wise to step-wise, enabling the policy model to achieve superior reasoning capabilities.

We conduct experiments on both in-domain and out-of-domain mathematical reasoning datasets with four widely used backbone LLMs.
Experimental results demonstrate that our Full-Step DPO consistently outperforms the DPO and Step-DPO baselines, validating its effectiveness in enhancing reasoning performance.
Our contributions can be summarized as follows:
\begin{itemize}[leftmargin=*]
    \setlength\itemsep{0.5em}
    \item
        We propose the Full-Step-DPO framework with a novel step-wise DPO loss that dynamically adjusts each step's gradient based on its reward, enabling step-wise optimization rather than solution-wise and enhancing reasoning ability.
    \item
        We train a self-supervised PRM to provide step-wise rewards for preference learning and explore a more efficient approach for automatically constructing PRM training data.       
    \item
        Extensive experiments on widely used mathematical benchmarks and base language models showcase the remarkable effectiveness of our method.
\end{itemize}

\section{Related Work}
\paragraph{Mathematical Reasoning}
Mathematical reasoning task is one of the most challenging tasks for LLMs. Various approaches have been explored to improve or elicit the mathematical reasoning ability of LLMs. 
A number of approaches have either continually pre-trained the base model on a vast of mathematical datasets \cite{llemma, deepseekmath} or used supervised fine-tuning with substantial synthetic datasets distilled from cutting-edge models \cite{wizardmath, metamath, orcamath, chatglm}.
Another line of work focuses on enhancing test-time computation by generating multiple solutions, developing separate reward models at either the outcome or process level to rerank these solutions \cite{cobbe2021training, lightman2023let}, or employing decoding strategies guided by the reward model \cite{ovm, xie2024self, mathshepherd, wu2024enhancing}.
In addition, Reinforcement Learning’s potential in general domains, demonstrated by \citet{gpt4} and \citet{llama}, some studies have explored its use in mathematical reasoning \cite{mathshepherd, orcamath, smaug}.

\paragraph{Preference Learning}
Recently, preference learning \cite{kto, mao2024don, mao2024simple} has attracted significant attention due to its ability to align with human preferences and distinguish between positive and negative examples. While these methods, like DPO \cite{dpo}, have proven effective in general domains, it offers only marginal benefits for mathematical reasoning \cite{smaug}. Some works \cite{alistep, cuhkstep} suggest that DPO's focus on coarse solution-wise preferences makes it less effective at correcting errors in multi-step reasoning, hindering reasoning improvement. Therefore, Step-DPO \cite{cuhkstep} was proposed, which first identifies the first erroneous step, and then optimizes only this erroneous step along with the corresponding correct one. Although this approach enhances mathematical reasoning capabilities, it totally overlooks the other steps in long-chain reasoning, which also provide valuable information and should not be completely disregarded. Building on this consideration, we propose Full-Step-DPO, which fully accounts for each step by dynamically optimizing all steps in the reasoning process.

\paragraph{Step-wise Supervision}
Recent findings by \citet{lightman2023let} suggest that step-wise supervision outperforms outcome-wise, due to the provision of more detailed feedback. 
However, training a PRM requires either costly manual annotation \cite{lightman2023let} or significant computational resources \cite{grace, mathshepherd}, which hinders the advancement and practical application of PRM.
Therefore, in this paper, we aim to build a PRM for mathematical reasoning without relying on human annotation and with reduced computational resources. 
Additionally, we explore the effectiveness of the PRM in decoding and preference learning scenarios.

\section{Full-Step DPO}

In this section, we elaborate the proposed Full-Step DPO framework.
We begin by reviewing the background of previous DPO and Step-DPO.
Then we introduce the novel Step-wise DPO Loss which optimizes with step-wise rewards,
and the Process Reward Model which automatically generate these step-wise rewards.
Finally we outline the complete training pipeline of our Full-Step-DPO.

\subsection{Preliminary} \label{sec:pre}
\textbf{DPO.} 
Direct Preference Optimization (DPO) \cite{dpo} is one of the most popular preference optimization methods. Instead of learning an explicit reward model, DPO directly uses pair-wise preference data to optimize the policy model with an
equivalent optimization objective. Specifically, given an input prompt $x$, and a preference data pair ($y^w$, $y^l$), DPO aims to maximize the probability of the entire preferred solution $y^w$ and minimize that of the dispreferred solution $y^l$. The optimization objective of DPO is:
\begin{equation*}
\begin{aligned}
\mathcal{L}_{\text{DPO}}(\theta) & = -\mathbb{E}_{(x, y^w, y^l) \sim \mathcal{D}} \Big[ \log \sigma \Big( \\ & \beta \log \frac{\pi_{\theta}(y^w \mid x)}{\pi_{\text{ref}}(y^w \mid x)} - \beta \log \frac{\pi_{\theta}(y^l \mid x)}{\pi_{\text{ref}}(y^l \mid x)} \Big) \Big] 
\end{aligned}
\end{equation*}
where $\pi_{\theta}(\cdot|x)$ is the policy model to be optimized, $\pi_{\text{ref}}(\cdot|x)$ is the reference model, $(x, y^w, y^l)$ are preference pairs, $\sigma$ is the sigmoid function, $\beta$ is a parameter controlling the deviation from the reference model. \\

\noindent
\textbf{Step-DPO.} 
Although DPO performs well on chat benchmarks, it is less effective for long-chain reasoning tasks like mathematical problems. Step-DPO \cite{cuhkstep} attributes this to DPO's inability to consider the sequential nature of mathematical reasoning, as rejecting an entire dispreferred solution may inadvertently penalize correct preceding steps, introducing significant noise. 
To address this, Step-DPO optimizes only the first incorrect step. As shown in \Cref{fig:fig1}, given a math problem and a series of initial correct reasoning steps $\{s_1, ..., s_{k-1}\}$, Step-DPO aims to maximize the probability of the correct next step $s^w_k$ and minimize the probability of the incorrect one $s^l_k$. Note that $s^w_k$ and $s^l_k$ refer to single steps, not all subsequent steps. The loss function used is still the vanilla DPO loss.

\subsection{Step-wise DPO Loss}
\label{sec:stepdpoloss}
We now introduce the novel Step-wise DPO loss, which performs step-wise optimization using step-wise rewards.
Although the motivation behind Step-DPO is reasonable, focusing solely on optimizing the first erroneous step and neglecting the valuable information provided by other steps may not be optimal. Additionally, we contend that it is not truly a step-wise DPO, as it still relies on the standard solution-wise DPO loss and resembles more of a data construction method. 

To address this, we modify the vanilla DPO loss to the step-wise DPO loss, dynamically weighting the gradients of each step based on its reward, thereby enabling true step-wise optimization.
Let's start with the gradient of the loss function $\mathcal{L}_{\text{DPO}}$. 
The gradient with respect to the parameters $\theta$ can be written as:
\begin{equation*}
\begin{aligned}
 \nabla_\theta \mathcal{L} &\!\!=\!\! -\beta  \mathbb{E}_{(x, y^w, y^l) \sim \mathcal{D}} [ \sigma \left( \hat{r}_\theta(x, y^l) - \hat{r}_\theta(x, y^w) \right)\\
 &[ \nabla_\theta \log \pi_\theta(y^w \mid x) \!\!-\!\! \nabla_\theta\log {\pi_\theta(y^l \mid x)} ]]
\end{aligned}
\label{eq:eq3}
\end{equation*}
where $\hat{r}_\theta(x, y) = \beta \log \frac{\pi_{\theta}(y \mid x)}{\pi_{\text{ref}}(y \mid x)}$. Intuitively, the gradient indiscriminately increases the likelihood of whole $y^w$ and decreases the likelihood of whole $y^l$.
To achieve dynamically weighting, we break $\nabla_\theta \log \pi_\theta(y \mid x)$ into a step-wise form and weight the gradient as follows:
\begin{equation*}
\small
\begin{aligned}
 \nabla_\theta \mathcal{L} = &
  -\beta \mathbb{E}_{(x, y^w, y^l) \sim \mathcal{D}} \Bigg[ \sigma \left( \hat{r}_\theta(x, y^l) - \hat{r}_\theta(x, y^w) \right)\\
 & \Bigg[ \sum_{i=1}^{K^w} \alpha_i^w\nabla_\theta \log \pi_\theta (s_i^w \mid x,s_{:<i}^w) \\
 & - \sum_{i=1}^{K^l} \alpha_i^l\nabla_\theta \log \pi_\theta (s_i^l \mid x,s_{:<i}^l) \Bigg]
\Bigg]
\end{aligned}
\end{equation*}
where $s_i$ represents the $i$-th reasoning step of the solution $y$, $s_{:<i}$ denotes all reasoning steps preceding $s_i$ , $K$ is the total number of steps, and
$\alpha_i$ is the weight coefficient of $s_i$, calculated based on the reward of $s_i$ as shown below:
\begin{equation*}
\begin{aligned}
    \alpha_{i} =\begin{cases} 
    \frac{e^{\gamma r_{s_i}}}{\sum_j e^{\gamma r_{s_j}}} , & \ s_i \in y^w \\
    \frac{e^{-\gamma r_{s_i}}}{\sum_j e^{-\gamma r_{s_j}}} , & \ s_i \in y^l 
    \end{cases}
\end{aligned}
\label{eq:eq5}
\end{equation*}
where $r_{s_i}$ is the reward of the step $s_i$, which will be introduced in the next subsection, and $\gamma$ is the temperature of the Softmax operation.
It is important to note that the calculation of $\alpha_{i}$ differs between the preferred solution $y^w$ and the dispreferred solution $y^l$. 
For preferred solutions, a higher reward indicates a greater likelihood of correct reasoning in that step, so the model should perform gradient ascent with greater intensity. Conversely, for dispreferred solutions, a lower reward suggests a higher chance of incorrect reasoning, and the model should apply gradient descent with greater intensity accordingly. 
This approach allows us to leverage all steps and adaptively adjust the weight of each step based on its probability of correctness, achieving true step-wise optimization.

Compared to Step-DPO methods that focus solely on a single step, our method optimizes all steps simultaneously, enabling better global optimization.
Noted that as $\gamma \!\! \to \!\! 0$, all steps will have equal weights, making Full-Step-DPO equivalent to vanilla DPO.

\subsection{Process Reward Models}
\label{sec:prm}
To obtain step-wise rewards, we train a Process Reward Model (PRM).
The biggest challenge in training a PRM is constructing a process supervision dataset. Previous studies \cite{uesato2022solving, lightman2023let} utilize human annotators to obtain step-wise labels, which requires advanced annotator skills and is quite costly. Later, MathShepherd \cite{mathshepherd} proposes using Monte Carlo estimation \cite{mcts} to automatically gather step-wise supervision, but it remained computationally expensive. In this section, we first examine the principles of Monte Carlo estimation, then present our simplified solution that significantly improves the efficiency of data construction. \\

\noindent
\textbf{Monte Carlo estimation.}
This approach assumes that the gold label $y_{s_i}$ of a step $s_i$ can be defined as the probability to deduce the correct answer $a^*$, and it includes both sampling and simulation phase. Specifically, given a math problem, it first randomly samples $M$ solutions, with each solution consisting of $K$ reasoning steps $\left \{s_{1}, s_{2}, \dots, s_{K} \right \}$, and $a$ represents the decoded answer from the last step $s_K$.
Then, to estimate the quality of reasoning step $s_i$ in a given solution, it simulates $N$ subsequent reasoning processes from this step: $\left \{ (s_{i+1,j}, \dots, s_{K,j}) \right \}_{j=1}^N$.
The golden label for $s_i$ is calculated as follows:
\begin{equation*}
    y_{s_{i}} = \frac{\sum_{j=1}^N\mathbb{I}(a_{j} = a^*)}{N} 
\label{eq:eq6}
\end{equation*}
where $a_j$ is the decoded answer for the $j$-th simulated solution, and $\mathbb{I}$ is the indicator function that returns 1 if $a_j=a^*$ and 0 otherwise. 
This two-stage approach is highly time-consuming, as it requires $N$ simulations for each of $K$ step across all $M$ solutions, resulting in a time complexity of $O(MNK)$. \\

\noindent
\textbf{Our efficient approach.}
It is important to note that there is a trade-off between the sampling number $M$ and the simulation number $N$ when computational resources are limited. A larger $M$ can provide more data for training the PRM, while a larger $N$ can result in higher accuracy of the labels $y_i$. 
In this paper, we found that the trained PRM performs reasonably well even with $N=1$ when $M$ is large, such as 32. This is likely because a larger $M$ introduces more diversity into the training data, making the PRM more tolerant to slight reductions in data precision caused by the limited simulation number.
This setting simplifies the PRM data construction by requiring only the sampling of $M$ solutions without the need for simulation, significantly reducing computational resources and lowering the time complexity to $O(M)$.
As a result, the gold label for step $s_i$ can be simplified as follows:
\begin{equation*}
\begin{aligned}
    y_{s_{i}} = \begin{cases} 
    1 & \text{if}\ a = a^* \\
    0 & \text{otherwise} 
    \end{cases}
\end{aligned}
\label{eq:eq7}
\end{equation*}
then the PRM could be trained as shown below:
\begin{equation*}
\begin{aligned}
\mathcal{L}_{\text{PRM}} = -\sum_{i=1}^{K} [y_{s_{i}} \log r_{s_{i}} + (1 - y_{s_{i}}) \log (1 - r_{s_{i}})]
\end{aligned}
\end{equation*}
where $y_{s_i}$ is the golden label for $s_i$, $r_{s_i}$ is the sigmoid score assigned by the PRM.
With the above PRM, we can automatically score each step in the reasoning chain, providing reward signals for the step-wise DPO loss and enabling step-wise optimization.

\begin{figure}[t]
\centering
\includegraphics[width=\linewidth]{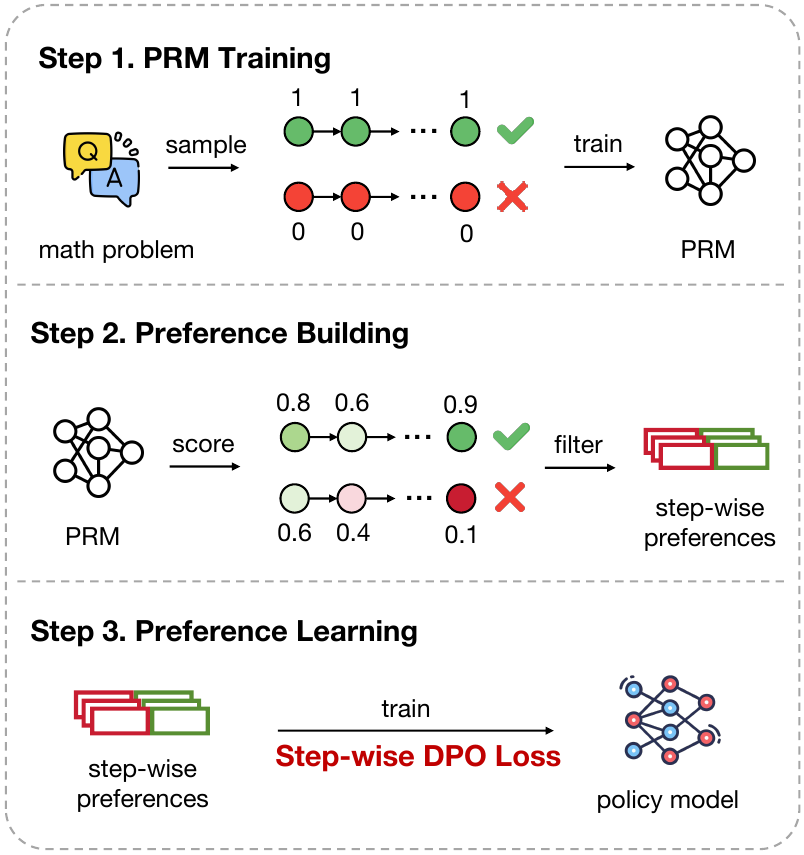}
\caption{The overall framework of Full-Step-DPO consists of three steps: (1) Training the PRM using the model itself and generated solutions. (2) Using the PRM to score and filter solutions to form preference data with step-wise rewards. (3) Training the policy model with the proposed step-wise DPO loss.}
\label{fig:fig2}
\end{figure}

\subsection{Training Pipeline}
\label{sec:pipeline}

Following previous methods \cite{mathshepherd, deepseekmath}, we adopt a standard training pipeline illustrated in \Cref{fig:fig2}: 
\begin{inparaenum}[(1)]
\item We begin by training a PRM with self-generated data, where higher reward values indicate a stronger likelihood of correct reasoning, while lower values suggest potential errors.
\item The trained PRM is then used to construct preference pairs with step-wise rewards. Specifically, we generate $M$ solutions for each math problem, score each step of these solutions with the PRM to produce a reward sequence, and calculate the average reward across all steps as the overall reward for each solution. We select the top $T$ correct solutions with the highest rewards and the bottom $T$ incorrect solutions with the lowest rewards to form $T^2$ step-wise preference pairs.
\item Finally, we update the policy model using the proposed step-wise DPO loss and the step-wise preference pairs, as described in Section \ref{sec:stepdpoloss}.
\end{inparaenum}

During the inference, a well-trained PRM can guide the decoding process and enhance the model's performance. Therefore, in addition to the standard greedy decoding, we explore three alternative decoding methods: 
\begin{inparaenum}[(1)]
\item Self-Consistency (SC) \cite{sc}: given a problem in the test set, we sample $K$ candidate solutions from the policy model. Instead of relying on the first decoded solution, we select the final answer based on majority voting over the answers provided by all sampled solutions. SC is a simple yet highly effective verification strategy.
\item Best-of-N (BoN): we similarly sample $K$ candidate solutions, score them using the reward model, and select the highest-scoring solution as the final answer. Following previous work \cite{lightman2023let, mathshepherd}, we use the minimum score across all steps as the final score assigned to a solution by the PRM. 
\item Step-wise Beam Search (SBS) \cite{ovm}: the PRM provides feedback at each step, offering more fine-grained guidance. Specifically, for each step, we first sample $b_1$ candidate subsequent steps, then score them using the PRM. The top $b_2$ steps are retained, and decoding continues until $b_2$ final solutions are reached. The detailed algorithm is provided in Appendix \ref{sec:sbs}. 
\end{inparaenum}

\section{Experiments}

\subsection{Experimental Setup} 

\textbf{Backbones.}  
To comprehensively validate the effectiveness of our proposed method, we adopt four popular open-source LLMs as the backbone models: MetaMath-Mistral-7B \cite{metamath}, Llama-3-8B \cite{llama}, DeepSeekMath-Base-7B \cite{deepseekmath} and Qwen2-7B \cite{qwen}. 
To improve these backbones' reasoning ability, Step-DPO \cite{cuhkstep} finetunes DeepSeekMath-Base-7B and Qwen2-7B on two open-source synthetic math datasets, MetaMath \cite{metamath} and MMIQC \cite{liu2024augmenting}, resulting in DeepSeekMath-Base-SFT~\footnote{\url{https://huggingface.co/xinlai/DeepSeekMath-Base-SFT}} and Qwen2-7B-SFT~\footnote{\url{https://huggingface.co/xinlai/Qwen2-7B-SFT}}, which greatly outperform their previous versions.
Following Step-DPO, we further finetune Llama3-8B to produce Llama3-8B-SFT. MetaMath-Mistral-7B has already been finetuned on MetaMath, so no additional finetuning was performed. \\

\noindent
\textbf{Baselines.}
For closed-source baselines, we compare our approach with OpenAI’s GPT-3.5 and GPT-4 \cite{gpt4}. We also benchmarked our method against recent high-performing mathematical LLMs, including WizardMath \cite{wizardmath}, MetaMath \cite{metamath}, InternLM-Math-7B \cite{internlm}, Qwen2-7B-Instruct\cite{qwen}, DeepSeekMath-Instruct \cite{deepseekmath}, InternLM-Math-20B \cite{internlm}, and Llama-3-70B-Instruct \cite{llama}.

Additionally, we compare it against DPO \cite{dpo} and Step-DPO \cite{cuhkstep}. Among these, \citet{cuhkstep} publicly release DeepSeekMath-Base-SFT-Step-DPO~\footnote{\url{https://huggingface.co/xinlai/DeepSeekMath-Base-SFT-Step-DPO}} and Qwen2-7B-SFT-Step-DPO~\footnote{\url{https://huggingface.co/xinlai/Qwen2-7B-SFT-Step-DPO}}, which we directly used for evaluation. Additionally, we trained MetaMath-Mistral-7B-Step-DPO and Llama-3-8B-SFT-Step-DPO using their publicly available code and dataset. \\

\noindent
\textbf{Datasets.} 
To ensure a fair comparison, we use the same training dataset~\footnote{\url{https://huggingface.co/datasets/xinlai/Math-Step-DPO-10K}} provided by Step-DPO \cite{cuhkstep}, which is synthesized from the training set of GSM8K \cite{gsm8k} and MATH \cite{math}. 
Noted that we only use the problem prompts in this dataset and do not use the step labels marked by GPT-4.

For in-domain evaluation, we conduct experiments on GSM8K and MATH, which contain 1,319 and 5,000 test problems, respectively. 
We also evaluate on two more challenging out-of-domain (OOD) test sets OCWCourses (OCW) \cite{ocw} and GaoKao2023 (GK2023) \cite{gaokao2023}. OCW contains of 272 undergraduate-level STEM problems requiring multi-step reasoning for most questions, while GK2023 includes 385 mathematics problems from the 2023 Chinese higher education entrance exam, translated into English. 
Accuracy serves as the evaluation metric. \\

\noindent
\textbf{Implementation Details.} 
During PRM training, we first randomly sample $M=32$ solutions for each math problem using Qwen2-7B-SFT and then label them as described in Section \ref{sec:prm}, resulting in the PRM training set.
Then, we add a classification-head to Qwen2-7B-SFT and train it on the PRM training set for one epoch. 
The batch size is 256, and the learning rate is 5e-7. 

To build preference learning datasets, we first sample $M = 32$ solutions for each math problem. The trained PRM then scores each solution, and we select $T=4$ solutions with the highest average rewards and $T=4$ with the lowest average rewards to randomly form 16 preference pairs.

During preference learning, the batch size is 64, the learning rate is 5e-7, $\beta$ is 0.05, and the reward temperature $\gamma$ is 0.5. 
We use the AdamW \cite{loshchilov2017decoupled} optimizer with a linear decay learning rate scheduler and only train one epoch. The warm-up ratio is 0.05.

During the decoding phase, we conduct experiments with two settings for SC and BoN, using $K = 5$ and $K = 15$. For Step-wise Beam Search, to ensure fair comparison, we test two configurations: $b_1 = 5, b_2 = 1$ (corresponding to $K=5$) and $b_1 = 5, b_2 = 3$ (corresponding to $K=15$). The sampling temperature is set to 0.8.

All the experiments are conducted on a server equipped with 8 NVIDIA A100-80GB GPUs and 512GB of system RAM. 
The implementation frameworks are PyTorch \cite{paszke2017automatic}, DeepSpeed \cite{rasley2020deepspeed}, and Huggingface \cite{wolf2019huggingface}.

\subsection{Main Results}
Table \ref{tab:tab1} provides a comprehensive comparison of various models on both MATH and GSM8K, including open-source and closed-source LLMs. We find that:
\begin{inparaenum}[(1)]
    \item Consistent with previous studies \cite{smaug}, DPO exhibits notable instability. Its performance shows slight degradation on MetaMath-Mistral-7B and MetaMath-Mistral-7B-SFT backbones, while the accuracy drops sharply to around 20\% on Qwen2-7B-SFT. It achieves a slight performance improvement only when applied to the DeepSeekMath-Base-SFT.
    \item Step-DPO achieves only minimal improvements across all backbones, with gains generally around 1\% and, in some settings, even slight performance drops. We evaluate the publicly released Step-DPO model using its official script, and the results may differ slightly from those reported in the Step-DPO paper. Similar issues have also been observed by other researchers~\footnote{\url{https://github.com/dvlab-research/Step-DPO/issues/2}}. 
    \item Our Full-Step-DPO consistently outperforms Step-DPO across all backbones. Specifically, when applied to MetaMath-Mistral-7B and Llama-3-8B-SFT, our model achieves improvements of approximately 2.3\% to 3.7\%, while applied to the stronger backbones, DeepSeekMath-Base-SFT and Qwen2-7B-SFT, our method still delivers gains exceeding 1\%.
\end{inparaenum}
These results clearly demonstrate the effectiveness of our proposed approach, which considers all steps in the reasoning process rather than focusing on solution-wise preferences or only a single step. A case study can be found in Appendix \ref{sec:case_study}.

\begin{table}[t]
    \centering
    \renewcommand{\arraystretch}{1.0}
    \resizebox{\linewidth}{!}{
    \begin{tabular}{l|c|c}
        \toprule
        \textbf{Model} & \textbf{MATH (\%)} & \textbf{GSM8K (\%)} \\
        \midrule
        GPT-3.5 & 34.1 & 80.8 \\
        GPT-4 & 53.6 & 93.6 \\
        \midrule
        WizardMath & 10.7 & 54.9 \\
        MetaMath & 19.8 & 66.5 \\
        InternLM-Math-7B & 34.6 & 78.1 \\
        Qwen2-7B-Instruct & 49.6 & 82.3 \\
        DeepSeekMath-Instruct & 46.8 & 82.9 \\
        InternLM-Math-20B & 37.7 & 82.6 \\
        Llama-3-70B-Instruct & 50.4 & 93.0 \\
        \midrule
        MetaMath-Mistral-7B & 28.2 & 77.7 \\
        \hspace{1em}+ DPO & 24.8 \textcolor{downred}{\raisebox{0.5ex}{\footnotesize -3.4}} & 70.7 \textcolor{downred}{\raisebox{0.5ex}{\footnotesize -7.0}}  \\
        \hspace{1em}+ Step-DPO  & 28.9 \textcolor{upgreen}{\raisebox{0.5ex}{\footnotesize +0.7}} &  79.6 \textcolor{upgreen}{\raisebox{0.5ex}{\footnotesize +1.9}} \\
        \rowcolor{lightpurple}
        \hspace{1em}+ Full-Step-DPO & \textbf{30.5} \textcolor{upgreen}{\raisebox{0.5ex}{\footnotesize +2.3}} & \textbf{81.4} \textcolor{upgreen}{\raisebox{0.5ex}{\footnotesize +3.7}} \\
        \midrule
        Llama-3-8B-SFT & 32.6 & 78.5 \\
        \hspace{1em}+ DPO & 23.4 \textcolor{downred}{\raisebox{0.5ex}{\footnotesize -9.2}} & 62.3 \textcolor{downred}{\raisebox{0.5ex}{\footnotesize -16.2}} \\
        \hspace{1em}+ Step-DPO& 31.8 \textcolor{downred}{\raisebox{0.5ex}{\footnotesize -0.8}} & 80.1 \textcolor{upgreen}{\raisebox{0.5ex}{\footnotesize +1.6}} \\
        \rowcolor{lightpurple}
        \hspace{1em}+ Full-Step-DPO & \textbf{35.0} \textcolor{upgreen}{\raisebox{0.5ex}{\footnotesize +2.4}} & \textbf{82.0} \textcolor{upgreen}{\raisebox{0.5ex}{\footnotesize +3.5}}\\
        \midrule
        DeepSeekMath-Base-SFT & 51.7 & 86.4 \\
        \hspace{1em}+ DPO & 51.7 \textcolor{downred}{\raisebox{0.5ex}{\footnotesize -0}} & 87.3 \textcolor{upgreen}{\raisebox{0.5ex}{\footnotesize +0.9}} \\
        \hspace{1em}+ Step-DPO & 52.9 \textcolor{upgreen}{\raisebox{0.5ex}{\footnotesize +1.2}} & 86.6 \textcolor{upgreen}{\raisebox{0.5ex}{\footnotesize +0.2}} \\
        \rowcolor{lightpurple}
        \hspace{1em}+ Full-Step-DPO & \textbf{53.2} \textcolor{upgreen}{\raisebox{0.5ex}{\footnotesize +1.5}} & \textbf{87.9} \textcolor{upgreen}{\raisebox{0.5ex}{\footnotesize +1.5}} \\
        \midrule
        Qwen2-7B-SFT & 53.9 & 88.3 \\
        \hspace{1em}+ DPO & 20.0 \textcolor{downred}{\raisebox{0.5ex}{\footnotesize -23.9}} & 27.3 \textcolor{downred}{\raisebox{0.5ex}{\footnotesize -61.0}} \\
        \hspace{1em}+ Step-DPO & 54.9 \textcolor{upgreen}{\raisebox{0.5ex}{\footnotesize +1.0}} & 88.4 \textcolor{upgreen}{\raisebox{0.5ex}{\footnotesize +0.1}} \\
        \rowcolor{lightpurple}
        \hspace{1em}+ Full-Step-DPO & \textbf{55.4} \textcolor{upgreen}{\raisebox{0.5ex}{\footnotesize +1.5}} & \textbf{89.3} \textcolor{upgreen}{\raisebox{0.5ex}{\footnotesize +1.0}} \\
        \bottomrule
    \end{tabular}
    }
    \caption{Performance comparison of various models on MATH and GSM8K with greedy decoding.}
    \label{tab:tab1}
\end{table}

\begin{table}[t]
    \centering
    \resizebox{\linewidth}{!}{
    \renewcommand{\arraystretch}{1.0}
    \begin{tabular}{l|c|c}
        \toprule
        \textbf{Model} & \textbf{GK2023 (\%)} & \textbf{OCW (\%)} \\
        \midrule
        MetaMath-Mistral-7B & 15.8 & \textbf{10.7} \\
        \hspace{1em}+ DPO & 15.8 \textcolor{downred}{\raisebox{0.5ex}{\footnotesize -0}} & 7.7 \textcolor{downred}{\raisebox{0.5ex}{\footnotesize -3.0}} \\
        \hspace{1em}+ Step-DPO & 15.1 \textcolor{downred}{\raisebox{0.5ex}{\footnotesize -0.7}} & 7.0 \textcolor{downred}{\raisebox{0.5ex}{\footnotesize -3.7}}\\
        \rowcolor{lightpurple}
        \hspace{1em}+ Full-Step-DPO & \textbf{20.5} \textcolor{upgreen}{\raisebox{0.5ex}{\footnotesize +4.7}} & 9.9 \textcolor{downred}{\raisebox{0.5ex}{\footnotesize -0.8}} \\
        \midrule
        Llama-3-8B-SFT & 20.5 & 12.5 \\
        \hspace{1em}+ DPO & 11.7 \textcolor{downred}{\raisebox{0.5ex}{\footnotesize -8.8}} & 9.9 \textcolor{downred}{\raisebox{0.5ex}{\footnotesize -2.6}} \\
        \hspace{1em}+ Step-DPO & 19.7 \textcolor{downred}{\raisebox{0.5ex}{\footnotesize -0.8}} & 13.6 \textcolor{upgreen}{\raisebox{0.5ex}{\footnotesize +1.1}} \\
        \rowcolor{lightpurple}
        \hspace{1em}+ Full-Step-DPO & \textbf{22.1} \textcolor{upgreen}{\raisebox{0.5ex}{\footnotesize +1.6}} & \textbf{15.1} \textcolor{upgreen}{\raisebox{0.5ex}{\footnotesize +2.6}} \\
        \midrule
        DeepSeekMath-Base-SFT & 30.4 & 19.1 \\
        \hspace{1em}+ DPO & 31.2 \textcolor{upgreen}{\raisebox{0.5ex}{\footnotesize +0.8}}  & 18.4 \textcolor{downred}{\raisebox{0.5ex}{\footnotesize -0.7}} \\
        \hspace{1em}+ Step-DPO & 31.2 \textcolor{upgreen}{\raisebox{0.5ex}{\footnotesize +0.8}} & 18.0 \textcolor{downred}{\raisebox{0.5ex}{\footnotesize -1.1}}\\
        \rowcolor{lightpurple}
        \hspace{1em}+ Full-Step-DPO & \textbf{31.7} \textcolor{upgreen}{\raisebox{0.5ex}{\footnotesize +1.3}} & \textbf{20.2} \textcolor{upgreen}{\raisebox{0.5ex}{\footnotesize +1.1}} \\
        \midrule
        Qwen2-7B-SFT & 33.0 & 15.8 \\
        \hspace{1em}+ DPO & 8.8 \textcolor{downred}{\raisebox{0.5ex}{\footnotesize -24.2}}  & 8.1 \textcolor{downred}{\raisebox{0.5ex}{\footnotesize -7.7}} \\
        \hspace{1em}+ Step-DPO & 32.5 \textcolor{downred}{\raisebox{0.5ex}{\footnotesize -0.5}} & 15.8 \textcolor{downred}{\raisebox{0.5ex}{\footnotesize -0}}\\
        \rowcolor{lightpurple}
        \hspace{1em}+ Full-Step-DPO & \textbf{33.5} \textcolor{upgreen}{\raisebox{0.5ex}{\footnotesize +0.5}} & \textbf{18.4} \textcolor{downred}{\raisebox{0.5ex}{\footnotesize +2.6}} \\
        \bottomrule
    \end{tabular}}
    \caption{Performance comparison on out-of-domain math problems.}
    \label{tab:tab2}
\end{table}

\begin{figure}[t]
\centering
\includegraphics[width=1\linewidth]{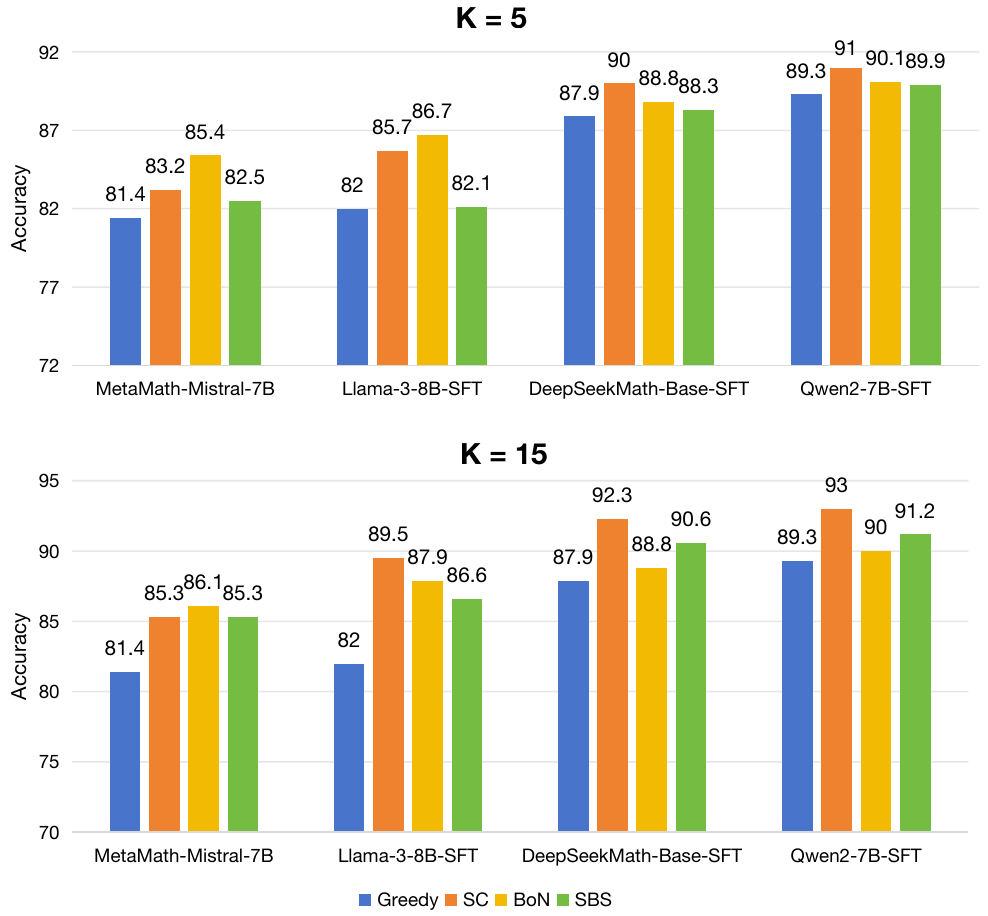}
\caption{Performance comparison of various verifications on GSM8K, with all models trained using our Full-Step-DPO.}
\label{fig:verify}
\end{figure}

\subsection{Results on OOD Datasets}
To further demonstrate the superiority of Full-Step-DPO, we evaluate the models on OOD datasets GK2023 and OCW, as shown in Table \ref{tab:tab2}.
On these competition-level math problems, DPO and Step-DPO often exhibit performance degradation under various settings, while our Full-Step-DPO consistently achieves performance improvements. 
The only exception occurs on the OCW dataset with MetaMath-Mistral-7B, where Full-Step-DPO shows a slight 0.8\% drop in accuracy. However, this drop is notably smaller than 3.0\% with DPO and the 3.7\% with Step-DPO. These results demonstrate the superior stability and resilience of Full-Step-DPO, particularly in handling challenging mathematical reasoning tasks. More experimental results on additional datasets can be found in Appendix \ref{sec:more_ood}.

\subsection{Results on Various Verification Strategies} 
\Cref{fig:verify} presents the performance of different verification strategies on GSM8K under two settings: $K=5$ and $K=15$. We find that: 
\begin{inparaenum}[(1)]
\item SC serves as a simple yet powerful validation method that significantly improves performance across all models. Even for the high-performing Qwen2-7B-SFT, which achieves an accuracy of 89.3\% with greedy decoding, SC further improves the accuracy to 93\% when the sampling size $K=15$. This result is already comparable to GPT-4's accuracy of 93.6\%.
\item Compared to SC, BoN often achieves further improvements on MetaMath-Mistral-7B and Llama-3-8B-SFT. However, on the highly capable DeepSeekMath-Base-SFT and Qwen2-7B-SFT, BoN underperforms SC, indicating that the benefits of the reward model diminish for very strong baseline models.
\item SBS performs worse than both SC and BoN across most settings, yet consistently surpasses Greedy decoding, aligning with findings from previous studies \cite{ovm, grace}. This may be because, during the early stages of inference, the reward model struggles to effectively distinguish the correctness of steps. 
\end{inparaenum}

\subsection{Analysis of PRMs}
\label{sec:ana_prms}

As discussed in \Cref{sec:prm}, the quality of the PRM may be influenced by the sampling number $M$ and simulation number $N$. To assess this, we conducted a controlled experiment with fixed $M=32$ and varying $N$. The trained PRM is then used to guide the decoding of MetaMath-Mistral-7B-Full-Step-DPO using BoN strategy with $K=15$.

\begin{figure}[t]
\centering
\includegraphics[width=\linewidth]{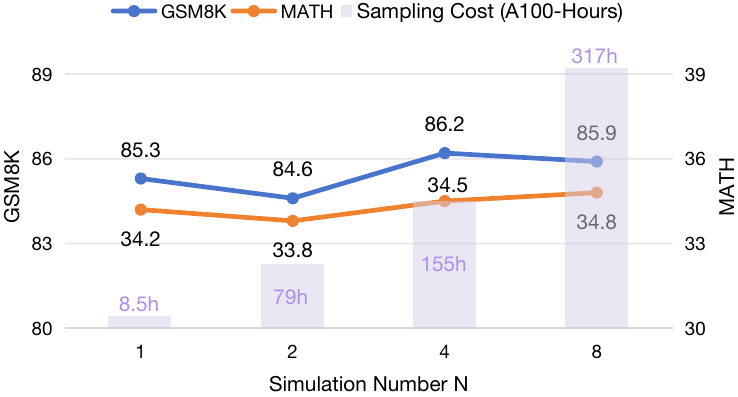}
\caption{Accuracy of MetaMath-Mistral-7B-Full-Step-DPO using BoN decoding with $K=15$. The PRM uses a fixed sampling number $M=32$, while the simulation number $N$ varies. Purple bars indicate the A100-Hours cost for constructing PRM training data.}
\label{fig:simulation}
\end{figure}

As shown in \Cref{fig:simulation}, the purple bars indicate the A100-Hours cost for constructing PRM training data, while blue and orange lines show GSM8K and MATH accuracy. When $N=1$, which represents our proposed method, the model achieves competitive performance with 85.3\% accuracy on GSM8K and 34.2\% on MATH, while only requiring 8.5h with a time complexity of $O(M)$. As $N$ increases, we observe that the performance fluctuation remains relatively minimal (within approximately 1\% range), while the sampling cost grows substantially with a complexity of $O(MNK)$. This empirical evidence suggests that our approach with $N=1$ achieves a good balance between sampling efficiency and model performance.
Additionally, we provide a comprehensive comparison between our PRM and other publicly available PRMs in \Cref{sec:diff_prms}.

\subsection{Sensitivity of Hyperparameters}

\begin{figure}[t]
\centering
\includegraphics[width=0.95\linewidth]{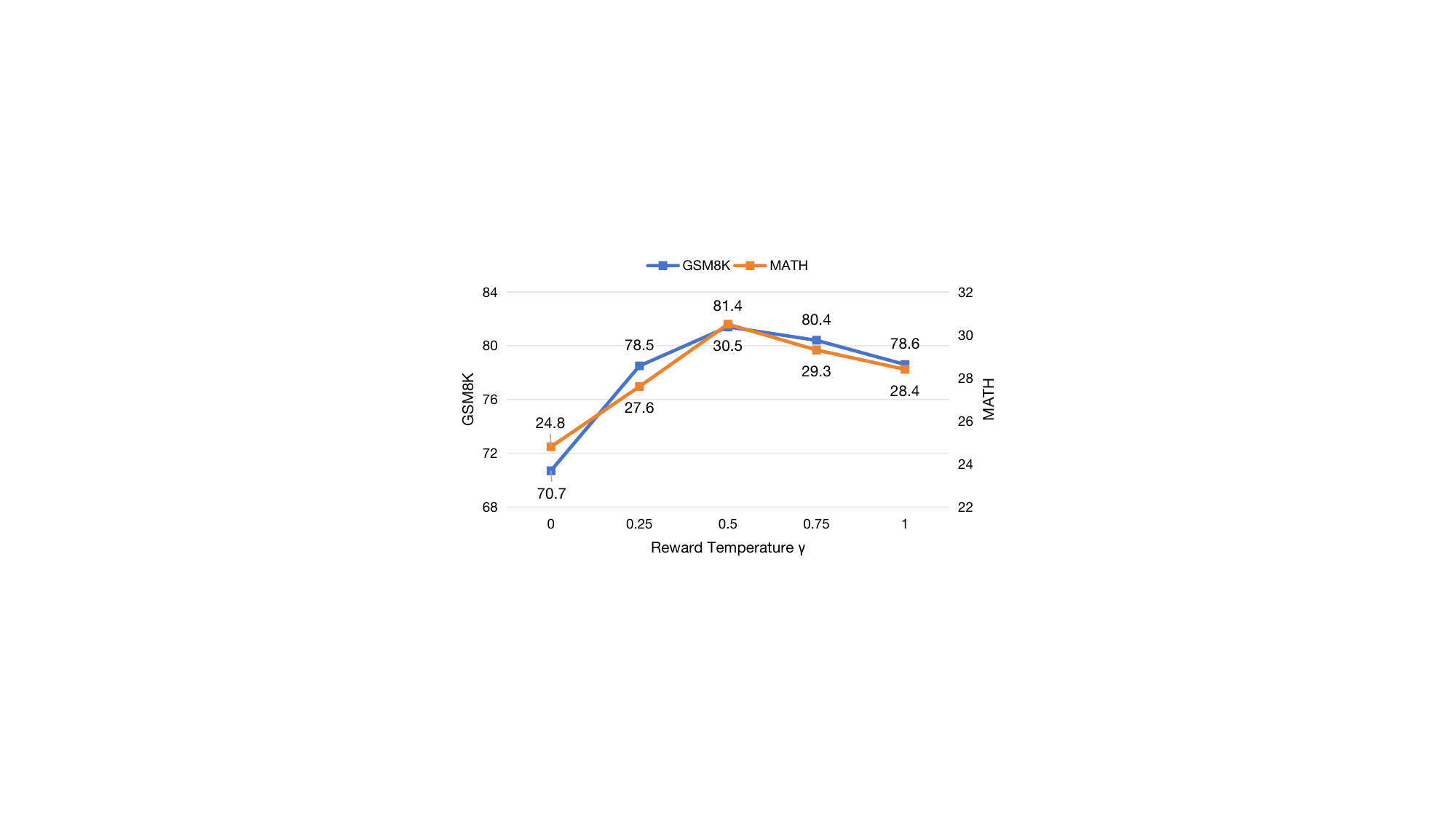}
\caption{Accuracy of MetaMath-Mistral-7B-Full-Step-DPO with different reward temperature $\gamma$.}
\label{fig:gamma}
\end{figure}

\noindent
In step-wise DPO loss, the reward temperature $\gamma$ reflects the level of trust in the PRM. As $\gamma$ increases, the PRM model has a greater impact on the gradients. When $\gamma \to 0$, it indicates complete distrust in the PRM model, assigning equal weight coefficient to all steps, degrading in vanilla DPO. Conversely, when $\gamma \to \infty$, the loss function optimizes only the single step with the maximum or minimum reward in the solution, similar to Step-DPO.
\Cref{fig:gamma} presents the accuracy of MetaMath-Mistral-7B-Full-Step-DPO with different $\gamma$ values. The experimental results indicate that introducing a PRM to weight the gradients indeed effectively enhances optimization efficiency and improves performance. Additionally, this experiment demonstrates that there is a sweet spot for the reward temperature $\gamma$; excessively high or low $\gamma$ will reduce accuracy.\\

\section{Conclusion}
In this work, we propose Full-Step-DPO, a novel framework for mathematical reasoning that optimizes each step in the entire reasoning chain using step-wise rewards. 
To achieve this, we train a self-supervised Process Reward Model to automatically score reasoning steps, eliminating reliance on external annotations. We also propose a novel Step-Wise DPO Loss that dynamically updates gradients based on the rewards for individual steps, enabling step-wise optimization and enhancing the reasoning ability of policy models.
Experimental results on various benchmarks validate the effectiveness of Full-Step-DPO, paving the way for its application to other reasoning-intensive tasks.

\section*{Limitations}
While we have conducted comprehensive experiments to demonstrate the effectiveness of Full-Step-DPO, several limitations remain. 
First, recent advancements suggest that generative reward models outperform the discriminative reward model used in this work. Exploring how generative reward models can further enhance mathematical reasoning capabilities would be a valuable direction for future research. 
Second, during preference data construction, the current strategy of selecting samples based on average reward is relatively simple. Investigating more advanced sample selection strategies may lead to further improvements. 
Finally, the step-wise DPO loss proposed in this paper is highly adaptable to other reasoning tasks, such as code generation. Conducting experiments on a broader range of tasks would provide additional evidence of the advantages of our method.

\section*{Acknowledgement}
This research/project is supported by the National Research Foundation, Singapore under its AI Singapore Programme (AISG Award No: AISG2-TC-2022-005).
We also wish to extend their heartfelt gratitude to the Sea AI Lab for their generous support in providing the necessary equipment and computational resources critical for the successful completion of this research.

\bibliography{custom}

\begin{thebibliography}{54}
\providecommand{\natexlab}[1]{#1}

\bibitem[{Achiam et~al.(2023)Achiam, Adler, Agarwal, Ahmad, Akkaya, Aleman, Almeida, Altenschmidt, Altman, Anadkat et~al.}]{gpt4}
Josh Achiam, Steven Adler, Sandhini Agarwal, Lama Ahmad, Ilge Akkaya, Florencia~Leoni Aleman, Diogo Almeida, Janko Altenschmidt, Sam Altman, Shyamal Anadkat, et~al. 2023.
\newblock \href {https://arxiv.org/abs/2303.08774} {Gpt-4 technical report}.
\newblock \emph{arXiv preprint arXiv:2303.08774}.

\bibitem[{Azerbayev et~al.(2023)Azerbayev, Schoelkopf, Paster, Santos, McAleer, Jiang, Deng, Biderman, and Welleck}]{llemma}
Zhangir Azerbayev, Hailey Schoelkopf, Keiran Paster, Marco~Dos Santos, Stephen McAleer, Albert~Q Jiang, Jia Deng, Stella Biderman, and Sean Welleck. 2023.
\newblock \href {https://arxiv.org/abs/2310.10631} {Llemma: An open language model for mathematics}.
\newblock \emph{arXiv preprint arXiv:2310.10631}.

\bibitem[{Bai et~al.(2023)Bai, Bai, Chu, Cui, Dang, Deng, Fan, Ge, Han, Huang et~al.}]{qwen}
Jinze Bai, Shuai Bai, Yunfei Chu, Zeyu Cui, Kai Dang, Xiaodong Deng, Yang Fan, Wenbin Ge, Yu~Han, Fei Huang, et~al. 2023.
\newblock \href {https://arxiv.org/abs/2309.16609} {Qwen technical report}.
\newblock \emph{arXiv preprint arXiv:2309.16609}.

\bibitem[{Chen et~al.(2024)Chen, Liao, Li, and Fan}]{alistep}
Guoxin Chen, Minpeng Liao, Chengxi Li, and Kai Fan. 2024.
\newblock \href {https://arxiv.org/abs/2406.10858} {Step-level value preference optimization for mathematical reasoning}.
\newblock \emph{arXiv preprint arXiv:2406.10858}.

\bibitem[{Cobbe et~al.(2021{\natexlab{a}})Cobbe, Kosaraju, Bavarian, Chen, Jun, Kaiser, Plappert, Tworek, Hilton, Nakano et~al.}]{cobbe2021training}
Karl Cobbe, Vineet Kosaraju, Mohammad Bavarian, Mark Chen, Heewoo Jun, Lukasz Kaiser, Matthias Plappert, Jerry Tworek, Jacob Hilton, Reiichiro Nakano, et~al. 2021{\natexlab{a}}.
\newblock \href {https://arxiv.org/abs/2110.14168} {Training verifiers to solve math word problems}.
\newblock \emph{arXiv preprint arXiv:2110.14168}.

\bibitem[{Cobbe et~al.(2021{\natexlab{b}})Cobbe, Kosaraju, Bavarian, Chen, Jun, Kaiser, Plappert, Tworek, Hilton, Nakano et~al.}]{gsm8k}
Karl Cobbe, Vineet Kosaraju, Mohammad Bavarian, Mark Chen, Heewoo Jun, Lukasz Kaiser, Matthias Plappert, Jerry Tworek, Jacob Hilton, Reiichiro Nakano, et~al. 2021{\natexlab{b}}.
\newblock \href {https://arxiv.org/abs/2110.14168} {Training verifiers to solve math word problems}.
\newblock \emph{arXiv preprint arXiv:2110.14168}.

\bibitem[{Coulom(2006)}]{mcts}
R{\'e}mi Coulom. 2006.
\newblock \href {https://www.semanticscholar.org/paper/Efficient-Selectivity-and-Backup-Operators-in-Tree-Coulom/02cc6a5944d57d2353a55639c7b77336b94f29b6} {Efficient selectivity and backup operators in monte-carlo tree search}.
\newblock In \emph{International conference on computers and games}, pages 72--83. Springer.

\bibitem[{Ethayarajh et~al.(2024)Ethayarajh, Xu, Muennighoff, Jurafsky, and Kiela}]{kto}
Kawin Ethayarajh, Winnie Xu, Niklas Muennighoff, Dan Jurafsky, and Douwe Kiela. 2024.
\newblock \href {https://arxiv.org/abs/2402.01306} {Kto: Model alignment as prospect theoretic optimization}.
\newblock \emph{arXiv preprint arXiv:2402.01306}.

\bibitem[{Hendrycks et~al.(2021)Hendrycks, Burns, Kadavath, Arora, Basart, Tang, Song, and Steinhardt}]{math}
Dan Hendrycks, Collin Burns, Saurav Kadavath, Akul Arora, Steven Basart, Eric Tang, Dawn Song, and Jacob Steinhardt. 2021.
\newblock \href {https://arxiv.org/abs/2103.03874} {Measuring mathematical problem solving with the math dataset}.
\newblock \emph{arXiv preprint arXiv:2103.03874}.

\bibitem[{Hosseini et~al.(2014)Hosseini, Hajishirzi, Etzioni, and Kushman}]{addsub}
Mohammad~Javad Hosseini, Hannaneh Hajishirzi, Oren Etzioni, and Nate Kushman. 2014.
\newblock \href {https://aclanthology.org/D14-1058/} {Learning to solve arithmetic word problems with verb categorization}.
\newblock In \emph{Proceedings of the 2014 Conference on Empirical Methods in Natural Language Processing (EMNLP)}, pages 523--533.

\bibitem[{Huang et~al.(2023)Huang, Chen, Mishra, Zheng, Yu, Song, and Zhou}]{huang2023large}
Jie Huang, Xinyun Chen, Swaroop Mishra, Huaixiu~Steven Zheng, Adams~Wei Yu, Xinying Song, and Denny Zhou. 2023.
\newblock \href {https://arxiv.org/abs/2310.01798} {Large language models cannot self-correct reasoning yet}.
\newblock \emph{arXiv preprint arXiv:2310.01798}.

\bibitem[{Jiao et~al.(2024)Jiao, Qin, Liu, Chen, and Joty}]{pdpo}
Fangkai Jiao, Chengwei Qin, Zhengyuan Liu, Nancy~F Chen, and Shafiq Joty. 2024.
\newblock \href {https://arxiv.org/abs/2402.00658} {Learning planning-based reasoning by trajectories collection and process reward synthesizing}.
\newblock \emph{arXiv preprint arXiv:2402.00658}.

\bibitem[{Kaddour et~al.(2023)Kaddour, Harris, Mozes, Bradley, Raileanu, and McHardy}]{kaddour2023challenges}
Jean Kaddour, Joshua Harris, Maximilian Mozes, Herbie Bradley, Roberta Raileanu, and Robert McHardy. 2023.
\newblock \href {https://arxiv.org/abs/2307.10169} {Challenges and applications of large language models}.
\newblock \emph{arXiv preprint arXiv:2307.10169}.

\bibitem[{Khalifa et~al.(2023)Khalifa, Logeswaran, Lee, Lee, and Wang}]{grace}
Muhammad Khalifa, Lajanugen Logeswaran, Moontae Lee, Honglak Lee, and Lu~Wang. 2023.
\newblock \href {https://arxiv.org/abs/2305.14934} {Grace: Discriminator-guided chain-of-thought reasoning}.
\newblock In \emph{Findings of the Association for Computational Linguistics: EMNLP 2023}, pages 15299--15328.

\bibitem[{Lai et~al.(2024)Lai, Tian, Chen, Yang, Peng, and Jia}]{cuhkstep}
Xin Lai, Zhuotao Tian, Yukang Chen, Senqiao Yang, Xiangru Peng, and Jiaya Jia. 2024.
\newblock \href {https://arxiv.org/abs/2406.18629} {Step-dpo: Step-wise preference optimization for long-chain reasoning of llms}.
\newblock \emph{arXiv preprint arXiv:2406.18629}.

\bibitem[{Lewkowycz et~al.(2022)Lewkowycz, Andreassen, Dohan, Dyer, Michalewski, Ramasesh, Slone, Anil, Schlag, Gutman-Solo et~al.}]{ocw}
Aitor Lewkowycz, Anders Andreassen, David Dohan, Ethan Dyer, Henryk Michalewski, Vinay Ramasesh, Ambrose Slone, Cem Anil, Imanol Schlag, Theo Gutman-Solo, et~al. 2022.
\newblock \href {https://arxiv.org/abs/2206.14858} {Solving quantitative reasoning problems with language models}.
\newblock \emph{Advances in Neural Information Processing Systems}, 35:3843--3857.

\bibitem[{Liao et~al.(2024)Liao, Luo, Li, Wu, and Fan}]{gaokao2023}
Minpeng Liao, Wei Luo, Chengxi Li, Jing Wu, and Kai Fan. 2024.
\newblock \href {https://arxiv.org/abs/2401.08190} {Mario: Math reasoning with code interpreter output--a reproducible pipeline}.
\newblock \emph{arXiv preprint arXiv:2401.08190}.

\bibitem[{Lightman et~al.(2023)Lightman, Kosaraju, Burda, Edwards, Baker, Lee, Leike, Schulman, Sutskever, and Cobbe}]{lightman2023let}
Hunter Lightman, Vineet Kosaraju, Yura Burda, Harri Edwards, Bowen Baker, Teddy Lee, Jan Leike, John Schulman, Ilya Sutskever, and Karl Cobbe. 2023.
\newblock \href {https://arxiv.org/abs/2305.20050} {Let's verify step by step}.
\newblock \emph{arXiv preprint arXiv:2305.20050}.

\bibitem[{Liu and Yao(2024)}]{liu2024augmenting}
Haoxiong Liu and Andrew Chi-Chih Yao. 2024.
\newblock \href {https://arxiv.org/abs/2401.09003} {Augmenting math word problems via iterative question composing}.
\newblock \emph{arXiv preprint arXiv:2401.09003}.

\bibitem[{Loshchilov and Hutter(2017)}]{loshchilov2017decoupled}
Ilya Loshchilov and Frank Hutter. 2017.
\newblock \href {https://arxiv.org/abs/1711.05101} {Decoupled weight decay regularization}.
\newblock \emph{arXiv preprint arXiv:1711.05101}.

\bibitem[{Lu et~al.(2024)Lu, Zhou, Wang, Ren, Shi, Pan, Zhan, and Li}]{stepcont}
Zimu Lu, Aojun Zhou, Ke~Wang, Houxing Ren, Weikang Shi, Junting Pan, Mingjie Zhan, and Hongsheng Li. 2024.
\newblock \href {https://arxiv.org/abs/2407.00782} {Step-controlled dpo: Leveraging stepwise error for enhanced mathematical reasoning}.
\newblock \emph{arXiv preprint arXiv:2407.00782}.

\bibitem[{Luo et~al.(2023)Luo, Sun, Xu, Zhao, Lou, Tao, Geng, Lin, Chen, and Zhang}]{wizardmath}
Haipeng Luo, Qingfeng Sun, Can Xu, Pu~Zhao, Jianguang Lou, Chongyang Tao, Xiubo Geng, Qingwei Lin, Shifeng Chen, and Dongmei Zhang. 2023.
\newblock \href {https://arxiv.org/abs/2310.10631} {Wizardmath: Empowering mathematical reasoning for large language models via reinforced evol-instruct}.
\newblock \emph{arXiv preprint arXiv:2308.09583}.

\bibitem[{Mao et~al.(2024{\natexlab{a}})Mao, Li, Xu, Zhang, Chen, and Luu}]{mao2024simple}
Xin Mao, Feng-Lin Li, Huimin Xu, Wei Zhang, Wang Chen, and Anh~Tuan Luu. 2024{\natexlab{a}}.
\newblock \href {https://arxiv.org/abs/2410.04834} {As simple as fine-tuning: Llm alignment via bidirectional negative feedback loss}.
\newblock \emph{arXiv preprint arXiv:2410.04834}.

\bibitem[{Mao et~al.(2024{\natexlab{b}})Mao, Li, Xu, Zhang, and Luu}]{mao2024don}
Xin Mao, Feng-Lin Li, Huimin Xu, Wei Zhang, and Anh~Tuan Luu. 2024{\natexlab{b}}.
\newblock \href {https://arxiv.org/abs/2402.16030} {Don't forget your reward values: Language model alignment via value-based calibration}.
\newblock \emph{arXiv preprint arXiv:2402.16030}.

\bibitem[{Miao et~al.(2021)Miao, Liang, and Su}]{asdiv}
Shen-Yun Miao, Chao-Chun Liang, and Keh-Yih Su. 2021.
\newblock \href {https://arxiv.org/abs/2106.15772} {A diverse corpus for evaluating and developing english math word problem solvers}.
\newblock \emph{arXiv preprint arXiv:2106.15772}.

\bibitem[{Mitra et~al.(2024)Mitra, Khanpour, Rosset, and Awadallah}]{orcamath}
Arindam Mitra, Hamed Khanpour, Corby Rosset, and Ahmed Awadallah. 2024.
\newblock \href {https://arxiv.org/abs/2402.14830} {Orca-math: Unlocking the potential of slms in grade school math}.
\newblock \emph{arXiv preprint arXiv:2402.14830}.

\bibitem[{Pal et~al.(2024)Pal, Karkhanis, Dooley, Roberts, Naidu, and White}]{smaug}
Arka Pal, Deep Karkhanis, Samuel Dooley, Manley Roberts, Siddartha Naidu, and Colin White. 2024.
\newblock \href {https://arxiv.org/abs/2402.13228} {Smaug: Fixing failure modes of preference optimisation with dpo-positive}.
\newblock \emph{arXiv preprint arXiv:2402.13228}.

\bibitem[{Pan et~al.(2024)Pan, Wu, Li, and Luu}]{pan2024fallacy}
Fengjun Pan, Xiaobao Wu, Zongrui Li, and Anh~Tuan Luu. 2024.
\newblock \href {https://doi.org/10.18653/v1/2024.emnlp-main.794} {Are {LLM}s good zero-shot fallacy classifiers?}
\newblock In \emph{Proceedings of the 2024 Conference on Empirical Methods in Natural Language Processing}, pages 14338--14364, Miami, Florida, USA. Association for Computational Linguistics.

\bibitem[{Paszke et~al.(2017)Paszke, Gross, Chintala, Chanan, Yang, DeVito, Lin, Desmaison, Antiga, and Lerer}]{paszke2017automatic}
Adam Paszke, Sam Gross, Soumith Chintala, Gregory Chanan, Edward Yang, Zachary DeVito, Zeming Lin, Alban Desmaison, Luca Antiga, and Adam Lerer. 2017.
\newblock \href {https://www.semanticscholar.org/paper/Automatic-differentiation-in-PyTorch-Paszke-Gross/b36a5bb1707bb9c70025294b3a310138aae8327a} {Automatic differentiation in pytorch}.
\newblock \emph{NIPS 2017 Workshop Autodiff Submission}.

\bibitem[{Patel et~al.(2021)Patel, Bhattamishra, and Goyal}]{svamp}
Arkil Patel, Satwik Bhattamishra, and Navin Goyal. 2021.
\newblock \href {https://arxiv.org/abs/2103.07191} {Are nlp models really able to solve simple math word problems?}
\newblock \emph{arXiv preprint arXiv:2103.07191}.

\bibitem[{Rafailov et~al.(2024)Rafailov, Sharma, Mitchell, Manning, Ermon, and Finn}]{dpo}
Rafael Rafailov, Archit Sharma, Eric Mitchell, Christopher~D Manning, Stefano Ermon, and Chelsea Finn. 2024.
\newblock \href {https://arxiv.org/abs/2305.18290} {Direct preference optimization: Your language model is secretly a reward model}.
\newblock \emph{Advances in Neural Information Processing Systems}, 36.

\bibitem[{Rasley et~al.(2020)Rasley, Rajbhandari, Ruwase, and He}]{rasley2020deepspeed}
Jeff Rasley, Samyam Rajbhandari, Olatunji Ruwase, and Yuxiong He. 2020.
\newblock \href {https://dl.acm.org/doi/10.1145/3394486.3406703} {Deepspeed: System optimizations enable training deep learning models with over 100 billion parameters}.
\newblock In \emph{Proceedings of the 26th ACM SIGKDD International Conference on Knowledge Discovery \& Data Mining}, pages 3505--3506.

\bibitem[{Shao et~al.(2024)Shao, Wang, Zhu, Xu, Song, Zhang, Li, Wu, and Guo}]{deepseekmath}
Zhihong Shao, Peiyi Wang, Qihao Zhu, Runxin Xu, Junxiao Song, Mingchuan Zhang, YK~Li, Y~Wu, and Daya Guo. 2024.
\newblock \href {https://arxiv.org/abs/2402.03300} {Deepseekmath: Pushing the limits of mathematical reasoning in open language models}.
\newblock \emph{arXiv preprint arXiv:2402.03300}.

\bibitem[{Shi et~al.(2023)Shi, Chen, Misra, Scales, Dohan, Chi, Sch{\"a}rli, and Zhou}]{gsmic}
Freda Shi, Xinyun Chen, Kanishka Misra, Nathan Scales, David Dohan, Ed~H Chi, Nathanael Sch{\"a}rli, and Denny Zhou. 2023.
\newblock \href {https://arxiv.org/abs/2302.00093} {Large language models can be easily distracted by irrelevant context}.
\newblock In \emph{International Conference on Machine Learning}, pages 31210--31227. PMLR.

\bibitem[{Song et~al.(2023)Song, Xiong, Zhu, Li, Wang, Tian, and Li}]{song2023restgpt}
Yifan Song, Weimin Xiong, Dawei Zhu, Cheng Li, Ke~Wang, Ye~Tian, and Sujian Li. 2023.
\newblock \href {https://arxiv.org/abs/2306.06624} {Restgpt: Connecting large language models with real-world applications via restful apis}.
\newblock \emph{arXiv preprint arXiv:2306.06624}.

\bibitem[{Touvron et~al.(2023)Touvron, Lavril, Izacard, Martinet, Lachaux, Lacroix, Rozi{\`e}re, Goyal, Hambro, Azhar et~al.}]{llama}
Hugo Touvron, Thibaut Lavril, Gautier Izacard, Xavier Martinet, Marie-Anne Lachaux, Timoth{\'e}e Lacroix, Baptiste Rozi{\`e}re, Naman Goyal, Eric Hambro, Faisal Azhar, et~al. 2023.
\newblock \href {https://arxiv.org/abs/2302.13971} {Llama: Open and efficient foundation language models}.
\newblock \emph{arXiv preprint arXiv:2302.13971}.

\bibitem[{Uesato et~al.(2022)Uesato, Kushman, Kumar, Song, Siegel, Wang, Creswell, Irving, and Higgins}]{uesato2022solving}
Jonathan Uesato, Nate Kushman, Ramana Kumar, Francis Song, Noah Siegel, Lisa Wang, Antonia Creswell, Geoffrey Irving, and Irina Higgins. 2022.
\newblock \href {https://arxiv.org/abs/2211.14275} {Solving math word problems with process-and outcome-based feedback}.
\newblock \emph{arXiv preprint arXiv:2211.14275}.

\bibitem[{Wang et~al.(2023{\natexlab{a}})Wang, Xie, Jiang, Mandlekar, Xiao, Zhu, Fan, and Anandkumar}]{wang2023voyager}
Guanzhi Wang, Yuqi Xie, Yunfan Jiang, Ajay Mandlekar, Chaowei Xiao, Yuke Zhu, Linxi Fan, and Anima Anandkumar. 2023{\natexlab{a}}.
\newblock \href {https://arxiv.org/abs/2305.16291} {Voyager: An open-ended embodied agent with large language models}.
\newblock \emph{arXiv preprint arXiv:2305.16291}.

\bibitem[{Wang et~al.(2023{\natexlab{b}})Wang, Li, Chen, Zhu, Lin, Cao, Liu, Liu, and Sui}]{wang2023large}
Peiyi Wang, Lei Li, Liang Chen, Dawei Zhu, Binghuai Lin, Yunbo Cao, Qi~Liu, Tianyu Liu, and Zhifang Sui. 2023{\natexlab{b}}.
\newblock \href {https://arxiv.org/abs/2305.17926} {Large language models are not fair evaluators}.
\newblock \emph{arXiv preprint arXiv:2305.17926}.

\bibitem[{Wang et~al.(2023{\natexlab{c}})Wang, Li, Shao, Xu, Dai, Li, Chen, Wu, and Sui}]{mathshepherd}
Peiyi Wang, Lei Li, Zhihong Shao, RX~Xu, Damai Dai, Yifei Li, Deli Chen, Y~Wu, and Zhifang Sui. 2023{\natexlab{c}}.
\newblock \href {https://arxiv.org/abs/2312.08935} {Math-shepherd: Verify and reinforce llms step-by-step without human annotations}.
\newblock \emph{CoRR, abs/2312.08935}.

\bibitem[{Wang et~al.(2022)Wang, Wei, Schuurmans, Le, Chi, Narang, Chowdhery, and Zhou}]{sc}
Xuezhi Wang, Jason Wei, Dale Schuurmans, Quoc Le, Ed~Chi, Sharan Narang, Aakanksha Chowdhery, and Denny Zhou. 2022.
\newblock \href {https://arxiv.org/abs/2203.11171} {Self-consistency improves chain of thought reasoning in language models}.

\bibitem[{Wolf et~al.(2019)Wolf, Debut, Sanh, Chaumond, Delangue, Moi, Cistac, Rault, Louf, Funtowicz et~al.}]{wolf2019huggingface}
Thomas Wolf, Lysandre Debut, Victor Sanh, Julien Chaumond, Clement Delangue, Anthony Moi, Pierric Cistac, Tim Rault, R{\'e}mi Louf, Morgan Funtowicz, et~al. 2019.
\newblock \href {https://arxiv.org/abs/1910.03771} {Huggingface's transformers: State-of-the-art natural language processing}.
\newblock \emph{arXiv preprint arXiv:1910.03771}.

\bibitem[{Wu et~al.(2024{\natexlab{a}})Wu, Nguyen, and Luu}]{wu2024survey}
Xiaobao Wu, Thong Nguyen, and Anh~Tuan Luu. 2024{\natexlab{a}}.
\newblock \href {https://doi.org/10.1007/s10462-023-10661-7} {A survey on neural topic models: Methods, applications, and challenges}.
\newblock \emph{Artificial Intelligence Review}.

\bibitem[{Wu et~al.(2024{\natexlab{b}})Wu, Nguyen, Zhang, Wang, and Luu}]{wu2024fastopic}
Xiaobao Wu, Thong~Thanh Nguyen, Delvin~Ce Zhang, William~Yang Wang, and Anh~Tuan Luu. 2024{\natexlab{b}}.
\newblock \href {https://openreview.net/forum?id=7t6aq0Fa9D} {{FAST}opic: Pretrained transformer is a fast, adaptive, stable, and transferable topic model}.
\newblock In \emph{The Thirty-eighth Annual Conference on Neural Information Processing Systems}.

\bibitem[{Wu et~al.(2024{\natexlab{c}})Wu, Pan, Wang, and Luu}]{wu2024akew}
Xiaobao Wu, Liangming Pan, William~Yang Wang, and Anh~Tuan Luu. 2024{\natexlab{c}}.
\newblock \href {https://doi.org/10.18653/v1/2024.emnlp-main.843} {{AKEW}: Assessing knowledge editing in the wild}.
\newblock In \emph{Proceedings of the 2024 Conference on Empirical Methods in Natural Language Processing}, pages 15118--15133, Miami, Florida, USA. Association for Computational Linguistics.

\bibitem[{Wu et~al.(2024{\natexlab{d}})Wu, Pan, Xie, Zhou, Zhao, Ma, Du, Mao, Luu, and Wang}]{wu2024antileak}
Xiaobao Wu, Liangming Pan, Yuxi Xie, Ruiwen Zhou, Shuai Zhao, Yubo Ma, Mingzhe Du, Rui Mao, Anh~Tuan Luu, and William~Yang Wang. 2024{\natexlab{d}}.
\newblock \href {https://arxiv.org/abs/2412.13670} {Antileak-bench: Preventing data contamination by automatically constructing benchmarks with updated real-world knowledge}.
\newblock \emph{arXiv preprint arXiv:2412.13670}.

\bibitem[{Wu et~al.(2024{\natexlab{e}})Wu, Zeng, Zhang, Tan, Shen, and Jiang}]{wu2024enhancing}
Zhenyu Wu, Qingkai Zeng, Zhihan Zhang, Zhaoxuan Tan, Chao Shen, and Meng Jiang. 2024{\natexlab{e}}.
\newblock \href {https://arxiv.org/abs/2410.12934} {Enhancing mathematical reasoning in llms by stepwise correction}.
\newblock \emph{arXiv preprint arXiv:2410.12934}.

\bibitem[{Xie et~al.(2024{\natexlab{a}})Xie, Goyal, Zheng, Kan, Lillicrap, Kawaguchi, and Shieh}]{nusstep}
Yuxi Xie, Anirudh Goyal, Wenyue Zheng, Min-Yen Kan, Timothy~P Lillicrap, Kenji Kawaguchi, and Michael Shieh. 2024{\natexlab{a}}.
\newblock \href {https://arxiv.org/abs/2405.00451} {Monte carlo tree search boosts reasoning via iterative preference learning}.
\newblock \emph{arXiv preprint arXiv:2405.00451}.

\bibitem[{Xie et~al.(2024{\natexlab{b}})Xie, Kawaguchi, Zhao, Zhao, Kan, He, and Xie}]{xie2024self}
Yuxi Xie, Kenji Kawaguchi, Yiran Zhao, James~Xu Zhao, Min-Yen Kan, Junxian He, and Michael Xie. 2024{\natexlab{b}}.
\newblock \href {https://arxiv.org/abs/2305.00633} {Self-evaluation guided beam search for reasoning}.
\newblock \emph{Advances in Neural Information Processing Systems}, 36.

\bibitem[{Xu et~al.(2024)Xu, Liu, Liu, Hou, Li, Zhang, Wang, Zeng, Du, Zhao et~al.}]{chatglm}
Yifan Xu, Xiao Liu, Xinghan Liu, Zhenyu Hou, Yueyan Li, Xiaohan Zhang, Zihan Wang, Aohan Zeng, Zhengxiao Du, Wenyi Zhao, et~al. 2024.
\newblock \href {https://arxiv.org/abs/2404.02893} {Chatglm-math: Improving math problem-solving in large language models with a self-critique pipeline}.
\newblock \emph{arXiv preprint arXiv:2404.02893}.

\bibitem[{Ying et~al.(2024)Ying, Zhang, Li, Zhou, Shao, Fei, Ma, Hong, Liu, Wang et~al.}]{internlm}
Huaiyuan Ying, Shuo Zhang, Linyang Li, Zhejian Zhou, Yunfan Shao, Zhaoye Fei, Yichuan Ma, Jiawei Hong, Kuikun Liu, Ziyi Wang, et~al. 2024.
\newblock \href {https://arxiv.org/abs/2402.06332} {Internlm-math: Open math large language models toward verifiable reasoning}.
\newblock \emph{arXiv preprint arXiv:2402.06332}.

\bibitem[{Yu et~al.(2023{\natexlab{a}})Yu, Gao, and Wang}]{ovm}
Fei Yu, Anningzhe Gao, and Benyou Wang. 2023{\natexlab{a}}.
\newblock \href {https://arxiv.org/abs/2311.09724} {Outcome-supervised verifiers for planning in mathematical reasoning}.
\newblock \emph{arXiv preprint arXiv:2311.09724}.

\bibitem[{Yu et~al.(2023{\natexlab{b}})Yu, Jiang, Shi, Yu, Liu, Zhang, Kwok, Li, Weller, and Liu}]{metamath}
Longhui Yu, Weisen Jiang, Han Shi, Jincheng Yu, Zhengying Liu, Yu~Zhang, James~T Kwok, Zhenguo Li, Adrian Weller, and Weiyang Liu. 2023{\natexlab{b}}.
\newblock \href {https://arxiv.org/abs/2309.12284} {Metamath: Bootstrap your own mathematical questions for large language models}.
\newblock \emph{arXiv preprint arXiv:2309.12284}.

\bibitem[{Zheng et~al.(2024)Zheng, Chiang, Sheng, Zhuang, Wu, Zhuang, Lin, Li, Li, Xing et~al.}]{zheng2024judging}
Lianmin Zheng, Wei-Lin Chiang, Ying Sheng, Siyuan Zhuang, Zhanghao Wu, Yonghao Zhuang, Zi~Lin, Zhuohan Li, Dacheng Li, Eric Xing, et~al. 2024.
\newblock \href {https://arxiv.org/abs/2306.05685} {Judging llm-as-a-judge with mt-bench and chatbot arena}.
\newblock \emph{Advances in Neural Information Processing Systems}, 36.

\end{thebibliography}

\clearpage

\appendix
\onecolumn

\section{Step-wise Beam Search}
\label{sec:sbs}

Unlike conventional beam search, which relies on token-level probabilities, our method integrates the reward model with an associated reranking criterion. This enables for step-wise beam search (SBS) \cite{ovm, alistep}, effectively selecting the preferred solution path in mathematical reasoning, while incurring a lower computational cost compared to Monte Carlo Tree Search.
Specifically, for each step $t$, suppose the sampling size is $b_1$, the policy model ${\pi_{\theta}}$ produces a set of candidate steps $\mathbb{S}^{(1:t+1)}=\{\mathcal{S}_i^{(1:t+1)}\}_{i=1}^{b_1}$, where $\mathcal{S}_i^{(1:t+1)} =\left[ s_i^1, ..., s_i^{t+1}\right]$ is the $i$-th partial solution up to step $t+1$. Given the PRM ${\pi_{r}}$ that can score each step, we select the top-scoring steps with beam size $b_2$.
The algorithm is detailed in Algorithm \ref{alg}.
By focusing on the quality of each reasoning step rather than just the final solution, our method enhances the overall reasoning capabilities of the model. 

\begin{algorithm}[htbp]
\caption{Step-wise Beam Search}
\label{alg}
\begin{algorithmic}[1]
\State \textbf{Input:} Math problem $q$, Sampling size $b_1$, Beam size $b_2$, Maximum step $C$
\State \textbf{Output:} Best solution for $q$
\State \textbf{Models:} Policy model ${\pi_{\theta}}$ and PRM ${\pi_{r}}$
\Function{StepLevelBeamSearch}{$q, b_1, b_2$,$C$}
    \State Initialize step sequences $\mathbb{S} \gets \{\}$
    \State Use $\pi_\theta$ to sample initial steps $\{s_1^1, \dots, s_{b_1}^1\}$
    \State Use $\pi_r$ to score all initial steps $\{r_1^1, \dots, r_{b_1}^1\}$
    \State Select top-$b_1$ steps and add to $\mathbb{S}$
    \State Set current step counter $t \gets 1$
    \While{$t < C$}
        \If {All sequences in $\mathbb{S}$ are complete}
            \State Break
        \EndIf
        \State $\mathbb{S}_{\text{new}} \gets \{\}$
        \State $\mathbb{R} \gets \{\}$;
        \For{each solution $\mathcal{S}^{(1:t)}$ in $\mathbb{S}$}
            \For{$i = 1$ to $b_1$}
                \State $\mathcal{S}_i^{(1:t+1)} = \pi_{\theta}(\mathcal{S}^{(1:t)}; q)$
                \State $r_i^{(1:t+1)} = \pi_r(\mathcal{S}_i^{(1:t+1)}; q)$
                \State $\mathbb{S}_{\text{new}} \gets \mathbb{S}_{\text{new}} + \{\mathcal{S}_i^{(1:t+1)}\}$
                \State $\mathbb{R} \gets \mathbb{R} + \{r_i^{(1:t+1)}\}$
            \EndFor
        \EndFor
        \State $\mathbb{S}_{\text{new}} \gets$ top-$b_2$ rewarded solutions in $(\mathbb{S}_{\text{new}}, \mathbb{R})$
        \State $\mathbb{S} \gets \mathbb{S}_{\text{new}}$
        \State $t \gets t + 1$;
    \EndWhile
    \State \Return solution with highest final reward in $\mathbb{S}$
\EndFunction
\end{algorithmic}
\end{algorithm}

\section{Additional Experiments}

\subsection{More OOD Datasets}
\label{sec:more_ood}
We evaluate our method on five additional OOD mathematical reasoning datasets as our testbed. As shown in Table \ref{tab:more_ood}, our Full-Step-DPO consistently improves performance across all datasets, demonstrating the effectiveness and generalization ability of our approach on OOD mathematical reasoning tasks.
\begin{itemize}
    \item
        SVAMP \cite{svamp} includes 1000 math questions of up to fourth grade difficulty. These questions can be solved by expressions requiring no more than two operators.
    \item
        AddSub \cite{addsub} contains 395 math questions that involve addition and subtraction operations.
    \item
        ASDiv \cite{asdiv} contains 2215 English math questions of different problem types. Each question provides the corresponding equation and answer.
    \item 
        GSM-IC2 and GSM-ICM \cite{gsmic} are mathematical reasoning datasets containing irrelevant conditions within the problem descriptions each consisting of 1000 problems. Problems in GSM-IC2 require two steps to solve, while problems in GSM-ICM require more than two steps to solve.
\end{itemize}

\begin{table}[htbp]
\centering
\renewcommand{\arraystretch}{1.0}
\resizebox{\textwidth}{!}{
\begin{tabular}{l|c|c|c|c|c|c}
    \toprule
    \textbf{Model} & \textbf{SVAMP (\%)} & \textbf{AddSub (\%)} & \textbf{ASDiv (\%)} & \textbf{GSM-IC2 (\%)} & \textbf{GSM-ICM (\%)} & \textbf{Average (\%)} \\
    \midrule
    MetaMath-Mistral-7B & 79.1 & 86.6 & 81.2 & \textbf{77.9} & \underline{76.5} & 80.3 \\
    \hspace{1em}+ DPO & 72.7 & 53.4 & 73.8 & 60.5 & 65.2 & 65.1 \\
    \hspace{1em}+ Step-DPO & \underline{80.3} & \underline{86.9} & \underline{82.7} & \underline{76.4} & 76.3 & \underline{80.5} \\
    \rowcolor{lightpurple}
    \hspace{1em}+ Full-Step-DPO & \textbf{81.7} & \textbf{88.6} & \textbf{83.9} & 75.9 & \textbf{76.6} & \textbf{81.3} \\
    \midrule
    Llama-3-8B-SFT & \underline{82.8} & 88.4 & 85.0 & 80.1 & 79.5 & 83.2 \\
    \hspace{1em}+ DPO & 72.2 & 60.0 & 74.2 & 64.1 & 66.3 & 67.4 \\
    \hspace{1em}+ Step-DPO & 81.7 & \underline{88.5} & \underline{85.3} & \underline{80.7} & \underline{81.2} & \underline{83.5} \\
    \rowcolor{lightpurple}
    \hspace{1em}+ Full-Step-DPO & \textbf{82.9} & \textbf{88.8} & \textbf{86.4} & \textbf{82.1} & \textbf{81.6} & \textbf{84.4} \\
    \midrule
    DeepSeekMath-Base-SFT & 84.2 & \textbf{87.6} & \underline{91.0} & 85.4 & 85.2 & 86.7 \\
    \hspace{1em}+ DPO & 85.1 & 86.6 & 90.6 & 85.2 & 85.0 & 86.5 \\
    \hspace{1em}+ Step-DPO & \underline{85.3} & 85.3 & 90.7 & \underline{85.9} & \textbf{86.2} & \underline{86.7} \\
    \rowcolor{lightpurple}
    \hspace{1em}+ Full-Step-DPO & \textbf{85.9} & \underline{86.6} & \textbf{91.2} & \textbf{87.8} & \underline{85.3} & \textbf{87.4} \\
    \midrule
    Qwen2-7B-SFT & \underline{88.7} & \textbf{92.7} & 91.6 & \underline{93.7} & 91.6 & \underline{91.7} \\
    \hspace{1em}+ DPO & 23.5 & 25.6 & 27.0 & 30.1 & 29.5 & 27.1 \\
    \hspace{1em}+ Step-DPO & 88.1 & 92.2 & \underline{91.8} & \textbf{93.8} & 91.9 & 91.6 \\
    \rowcolor{lightpurple}
    \hspace{1em}+ Full-Step-DPO & \textbf{89.5} & \textbf{93.1} & \textbf{92.4} & 93.6 & \textbf{92.7} & \textbf{92.3} \\
    \bottomrule
\end{tabular}
}
\caption{Performance comparison on five additional OOD math problems. The best and second-best results are highlighted in bold and underlined, respectively.}
\label{tab:more_ood}
\end{table}

\subsection{Performance Comparison of Different PRMs}
\label{sec:diff_prms}
We compare our PRM with Math-Shepherd-PRM-7B using MetaMath-Mistral-7B as the base model under the Best-of-8 decoding strategy. During decoding, we sampled eight responses with temperature = 0.7. For both PRMs, each response was scored by the minimum reward among its reasoning steps, and the response with the highest score was selected as the final output.

As shown in \Cref{tab:diff_prms}, our PRM achieves 84.3\% accuracy on GSM8K and 33.2\% on MATH, outperforming Math-Shepherd-PRM-7B (82.4\% on GSM8K, 32.6\% on MATH). While these results suggest the effectiveness of our approach, we acknowledge that the comparison may not be entirely comprehensive due to differences in backbone architecture and training data between the PRMs. Nevertheless, these findings, combined with our analysis in \Cref{sec:ana_prms}, demonstrate that our PRM can achieve competitive performance while significantly reducing the computational cost of PRM construction.

\begin{table}[htbp]
\centering
\resizebox{0.5\textwidth}{!}{
\begin{tabular}{lcc}
\toprule
\textbf{Model} & \textbf{MATH} & \textbf{GSM8K} \\
\midrule
MetaMath-Mistral-7B & 28.2 & 77.7 \\
\hspace{1em}+ Self-Consistency & 32.4 & 81.3 \\
\hspace{1em}+ Math-Shepherd-PRM-7B & 32.6 & 82.4 \\
\rowcolor{lightpurple}
\hspace{1em}+ Our-PRM-7B & \textbf{33.2} & \textbf{84.3}	\\
\bottomrule
\end{tabular}}
\caption{Performance comparison of different scoring strategies using Best-of-8 decoding on MATH and GSM8K benchmarks.}
\label{tab:diff_prms}
\end{table}

\subsection{Case Study}
\label{sec:case_study}
We conducted a detailed case study using a sample problem from the AddSub dataset, as shown in Table \ref{tab:case_study}. We compared outputs from DeepSeekMath-Base-SFT model and its Step-DPO and Full-Step-DPO variants. While both the base SFT model and Step-DPO model made incorrect reasoning by including the cost of unbought CDs, our Full-Step-DPO model generated correct reasoning steps. Our PRM scores accurately reflect the correctness of each reasoning step, with higher scores for correct steps and lower scores for incorrect ones.

\renewcommand{\arraystretch}{1.2}
\begin{table}[htbp]
\centering
\resizebox{\textwidth}{!}{
{\small
\begin{tabular}{l|p{12.5cm}}
\hline
\textbf{Question} & For his car, Mike spent \$118.54 on speakers and \$106.33 on new tires. Mike wanted 3 CD's for \$4.58 but decided not to. In total, how much did Mike spend on car parts? \\
\hline
\textbf{Gold Answer} & 224.87 \\
\hline
\multirow{3}{*}{\makecell[l]{\textbf{SFT}\\\textbf{Response}}} &
\textbf{Step 1}: Mike spent $118.54 on speakers and $106.33 on new tires. So the total cost of these two items is \$118.54 + \$106.33 = \$224.87.
\hfill \textcolor{upgreen}{\textbf{Score: 0.9348}} \\
\cline{2-2}
& \textbf{Step 2}: Mike wanted to buy 3 CD's for \$4.58 each, but he decided not to. So the total cost of the CD's he didn't buy is 3 * \$4.58 = \$13.74.
\hfill \textcolor{downred}{\textbf{Score: 0.3646}} \\
\cline{2-2}
& \textbf{Step 3}: To find out how much Mike spent in total on car parts, we need to add the total cost of the speakers and tires to the total cost of the CD's he didn't buy. So the total cost is \$224.87 + \$13.74 = \$238.61. The answer is: 238.61
\hfill \textcolor{downred}{\textbf{Score: 0.0427}} \\
\hline
\multirow{3}{*}{\makecell[l]{\textbf{Step-DPO}\\\textbf{Response}}} &
\textbf{Step 1}: First, Mike spent \$118.54 on speakers and \$106.33 on new tires. So the total cost of these two items is \$118.54 + \$106.33 = \$224.87.
\hfill \textcolor{upgreen}{\textbf{Score: 0.9406}} \\
\cline{2-2}
& \textbf{Step 2}: Mike also wanted 3 CD's, but he decided not to buy them. Each CD costs \$4.58, so the total cost of the CD's he didn't buy is 3 * \$4.58 = \$13.74.
\hfill \textcolor{downred}{\textbf{Score: 0.3036}} \\
\cline{2-2}
& \textbf{Step 3}: To find out how much Mike spent in total on car parts, we need to add the total cost of the speakers and tires to the total cost of the CD's he didn't buy. So, the total cost is \$224.87 + \$13.74 = \$238.61. The answer is: 238.61
\hfill \textcolor{downred}{\textbf{Score: 0.0445}} \\
\hline
\multirow{3}{*}{\makecell[l]{\textbf{Full-Step-DPO}\\\textbf{Response}}} &
\textbf{Step 1}: Mike spent \$118.54 on speakers and \$106.33 on new tires. So the total amount he spent on these two items is \$118.54 + \$106.33 = \$224.87.
\hfill \textcolor{upgreen}{\textbf{Score: 0.9313}} \\
\cline{2-2}
& \textbf{Step 2}: Mike wanted to buy 3 CD's for \$4.58 each, but he decided not to. So the total amount he would have spent on the CD's is 3 * \$4.58 = \$13.74.
\hfill \textcolor{upgreen}{\textbf{Score: 0.7408}} \\
\cline{2-2}
& \textbf{Step 3}: To find out how much Mike spent in total on car parts, we need to add the total amount he spent on speakers and tires to the amount he would have spent on CD's. So the total amount is \$224.87 + \$0 = \$224.87. The answer is: 224.87
\hfill \textcolor{upgreen}{\textbf{Score: 0.7363}} \\
\hline
\end{tabular}}}
\caption{Comparison of reasoning steps and their corresponding PRM scores across different models for AddSub dataset. The base model is DeepSeekMath-Base-SFT.}
\label{tab:case_study}
\end{table}

\end{document}